\documentclass[11pt,a4paper]{article}
\usepackage{amsmath,amssymb,amsthm}
\usepackage[margin=1in]{geometry}
\usepackage{hyperref}
\usepackage{natbib}
\usepackage{enumitem}
\usepackage{setspace}

\newtheorem{definition}{Definition}
\newtheorem{proposition}{Proposition}
\newtheorem{theorem}{Theorem}
\newtheorem{lemma}{Lemma}
\newtheorem{corollary}{Corollary}

\title{A Quantitative Definition of Intelligence}
\author{Kang-Sin Choi
 \\ \normalsize \textit{Ewha Womans University} \\
\normalsize \texttt{kangsin@ewha.ac.kr}
}
\date{}

\begin{document}

\maketitle

\begin{abstract}
We propose an operational, quantitative definition of intelligence for arbitrary physical systems. The intelligence density of a system is the ratio of the logarithm of its independent outputs to its total description length. A system memorizes if its description length grows with its output count; it knows if its description length remains fixed while its output count diverges. The criterion for knowing is generalization. A system knows its domain if a single finite mechanism can produce correct outputs across an unbounded range of inputs, rather than storing each answer individually. The definition places intelligence on a substrate-independent continuum from logic gates to brains. We then argue that meaning over a domain is a selection and ordering of functions that produces correct outputs where correctness is specifiable. We also define a measure of contextuality of an output as the inverse of its conditional Kolmogorov complexity given the context of prior outputs, which unifies correctness and independence into a single condition. Together, these refute Searle's third premise, that syntax is insufficient for semantics, over any domain where correctness is specifiable.
\end{abstract}

\section{Introduction}
\label{sec:introduction}

Turing \citeyearpar{turing1950} was the first to approach the question about intelligence with care (see also \citep{legg2007}). Rather than defining intelligence, he proposed replacing the question ``Can machines think?'' with a behavioral test. He anticipated virtually every objection raised in the following decades: the argument from consciousness, Lady Lovelace's objection that machines can only do what we tell them, and the argument from the continuity of the nervous system. He observed that machines ``take me by surprise with great frequency,'' that the substrate of computation, whether electrical, mechanical, or otherwise, ``cannot be of theoretical importance,'' and in a 1951 letter that the difference between machine and brain is ``largely a quantitative matter'' \citep{turing1951letter}.

Turing's intuitions were remarkably prescient, but he stopped short of a formal definition. The present paper supplies one. We propose a single quantitative metric that formalizes what Turing intuited, that intelligence is computation, and computation is measurable. The central distinction of this paper is between {\em memorizing} and {\em knowing}. A system memorizes if its description length grows with its output count; it knows if its description length remains fixed while its output count diverges. This distinction is not a matter of degree but of asymptotic structure, and it is the formal content of the intuition that intelligence is generalization. The central criterion is not the absolute value of the metric but its {\em scaling behavior}. A system knows its domain if, as the domain of inputs grows without bound, its intelligence density diverges rather than vanishes. A persistent source of confusion in this area has been the conflation of intelligence with consciousness \citep{seth2021}. We define intelligence only. Consciousness, whether it exists, what it requires, or whether machines can have it, is a separate question, beyond our scope.

In practice, intelligence is what gets tasks done correctly. Doing a task correctly requires the proper arrangement of knowledge, that is the right procedures chosen in the right order. This is function composition, which in the domains considered here coincides with what syntax amounts to. We show that over any domain where correctness can be specified, correct syntax is sufficient for correct outputs, and nothing beyond the arrangement is needed. We define the contextuality of an output as the inverse of its conditional complexity given the context of prior outputs, and show that high contextuality implies independence from unrelated outputs, so that a system meeting the knowing criterion performs the tasks of the domain rather than merely generating symbols. This answers Premise 3 of the Chinese Room debate, that syntax is insufficient for semantics, within any domain where correctness is specifiable, while Premise 2, concerning the origin of intentionality, is reframed and deferred to a companion treatment of understanding (Section~\ref{sec:intentionality}). The framework thereby provides a quantitative structure in which intelligence and meaning are two aspects of a single arrangement, one measuring the system and the other measuring what the system produces.

\section{The Chinese room revisited}

A natural starting point is Searle's Chinese Room Argument (CRA) \citep{searle1980}, which imagines a monolingual English speaker in a room, following a rulebook to manipulate Chinese symbols. Chinese questions are slipped in through a slot; following the book, the man produces Chinese answers that are indistinguishable from those of a native speaker. Searle contends that neither the man nor the system understands Chinese: the man merely shuffles symbols syntactically, without grasping their meaning.

The major replies all argue about {\em who} or {\em what} in the room understands, without defining what understanding {\em is}. The Systems Reply claims that ``the whole system'' (the man, the rulebook, the room) understands Chinese, but cannot say where in the system the understanding resides or how to measure it. Without a definition, Searle can always retreat to ``but that's not {\em real} understanding.'' His ultimate retreat is biological naturalism: understanding requires the specific causal powers of biological neurons \citep{searle1992}, a substrate claim, addressed by substrate independence (Section~\ref{sec:intelligence-density}). The common failure is not a lack of ingenuity but a lack of a shared criterion. We take a different approach. Rather than arguing about who in the room understands, we examine the rulebook itself and ask what it must contain. Our central claim is
\begin{equation}
\text{The rulebook must generalize.}
\end{equation}

In the original CRA, the rulebook is given and we never ask where it came from. However, the intelligence is {\em in the rulebook}. Someone (a programmer, a linguist, a culture) had to understand Chinese and the domains it covers in order to write it. The rules for handling arithmetic, geography, history, and open-ended conversation represent generalized knowledge. Once written, the book operates independently of its author, answering questions its author never considered. The book now contains the intelligence, produced by its author's understanding, but operating on its own. We do not say the student ``merely follows instructions.''

We now examine what this book must contain. A competent Chinese speaker can answer arithmetic questions posed in Chinese (``What is four times thirty-seven?'' or ``What gives $7,823 + 4,567$?'') expressed in Chinese numerals and characters. If the rulebook enables the room to pass the Turing test for Chinese, it must handle such questions for {\em arbitrary} numbers, since a Chinese interrogator can ask about any arithmetic problem. There are two possible designs for the arithmetic portion of the rulebook:
\begin{itemize}
\item[]\textbf{(Lookup table)} The book contains a table listing every possible arithmetic problem and its answer ($1 \times 1 = 1$, $2 \times 1 = 2$, \ldots, $34 \times 182 = 6{,}188$, \ldots), together with the corresponding Chinese characters.
\item[]\textbf{(Algorithm)} The book contains rules for multiplication via repeated addition, addition via carry rules, and carry rules via binary XOR and AND, together with rules for converting between Chinese numerals and the internal representation.

\end{itemize}

\begin{proposition} \label{prop:infinite}
The arithmetic portion of any finite rulebook that passes the Turing test for Chinese must be algorithmic, not a lookup table.
\end{proposition}
The set of arithmetic questions expressible in Chinese is therefore countably infinite. Every pair of natural numbers can be named, and each pair determines a distinct question. A lookup table for all such questions requires infinitely many entries, each of positive size, and hence requires infinite storage. The rulebook is a physical object with finite pages. Therefore, the arithmetic portion cannot be a lookup table. It must use a finite algorithm that generalizes across all inputs.

This argument uses arithmetic only because the infinity of the input domain is mathematically indisputable. But the same logic applies to every open-ended component of Chinese conversation, since there are infinitely many possible questions about geography, history, counting, comparisons, hypotheticals, and novel combinations of these. The arithmetic case is simply the cleanest proof that the rulebook must generalize. It follows that Searle's rulebook, if finite, necessarily contains algorithms that produce correct outputs for inputs never explicitly listed. That is, they go beyond what was ``told'' to the system. Whether this constitutes ``knowing'' depends on how one defines the term. We provide such a definition in Section~\ref{sec:intelligence-density}.

Choosing the algorithmic option introduces a computer. It is not important to whom it belongs, whether the man or the rulebook, but the computation has to be done somewhere. The computer and a set of rules for manipulating it, which we now call a {\em system}, generate the intelligence.\footnote{Our answer is not the Systems Reply. Wherever there is a finite mechanism that generalizes, an algorithm whose $\mathcal{I}(n)$ continues to grow as the domain scales, there is knowing. We do not claim the room ``as a whole'' knows; we claim the {\em rulebook} must know, because it must contain the computation that generalizes.}  One must also ask whether such a system can generate meaningful sentences, not merely syntactically well-formed ones. This question requires careful treatment; we defer it to Section~\ref{sec:arrangement}, where we argue that meaning over a domain is a selection and ordering of functions that produces correct outputs, and that a system which generalizes necessarily captures this structure.

The arguments of this section do not depend on a formal definition of intelligence. Proposition \ref{prop:infinite} requires only that the rulebook is finite and the domain is infinite. But a formal definition quantifies the insight. It measures {\em how much} a system generalizes, distinguishes generalization from memorization precisely, and places all systems on a common scale. We develop this definition in the next section.

\section{Intelligence density}
\label{sec:intelligence-density}

Intelligence need not be biological. Proposition \ref{prop:infinite} makes no reference to substrate: the argument that a finite rulebook must generalize applies whether the rulebook is paper, silicon, or neural tissue. Turing observed that Babbage's Analytical Engine was entirely mechanical: ``since all digital computers are in a sense equivalent, we see that this use of electricity cannot be of theoretical importance'' \citep{turing1950}. Any adequate definition of intelligence must therefore be substrate-independent: it must assign the same value to the same computation regardless of what physical medium implements it.

The above discussion of the CRA suggests a natural measure. The ratio of independent outputs to description length diverges for algorithms and vanishes for lookup tables. Taking the logarithm of the output count follows the information-theoretic convention, since independent contributions combine additively.
\begin{definition}[Intelligence density] \label{def:intelligence}
Let $S$ be a physical system with total description length $C(S)$, the length of the program and data required to specify $S$. Let $N(S)$ be the number of {\em independent} outputs $S$ can produce in response to distinct inputs. The intelligence density of $S$ is
\begin{equation}
\mathcal{I}(S) = \frac{\log N(S)}{C(S)}
\end{equation}
\end{definition}

Roughly, $\mathcal{I}$ measures how many genuinely different things a system can do per bit of its description. A system that produces many independent outputs from a short description has high intelligence density; one that requires a long description for each output has low intelligence density.

For a Turing machine, $C(S)$ is the contents of the tape before execution;\footnote{Working memory consumed during execution is not part of $C(S)$: the separation between the fixed program and the working space it uses is developed in Section~\ref{sec:iteration}.} for a neural network, the total size of its weights; for a lookup table, the number of entries times bits per entry. The independence condition below ensures that $N$ counts genuinely different outputs, so that the additive treatment of $\log N$ is meaningful. $\mathcal{I}$ is therefore a dimensionless ratio, independent of the logarithmic base provided $C$ is measured in the corresponding unit.\footnote{The worked examples in this paper adopt $\log_2$ and measure $C$ in bits, which fixes a particular numerical scale; the value of $\mathcal{I}$ is invariant under this choice.}

In fact, both $C(S)$ and $N(S)$ are {\em relative to a domain $D$}, arithmetic, language or chess. Also, they depend on {\em a chosen level of description}: a brain measured at the synaptic level with linguistic outputs is a different computational system from the same brain measured at the molecular level with protein-folding outputs (see Section \ref{sec:C}). The size parameter $n$ of the domain (digit count, board size, token count) is also implicit in this dependence. With this in mind, we use the shorthand $\mathcal{I}(S)$ with domain and $n$ implicit, writing $\mathcal{I}(n)$ or $\mathcal{I}(S, n)$ only when the scaling in $n$ is the point of interest. The relativity is not a defect of the measure but a correct reflection of the fact that intelligence is always intelligence {\em with respect to} some domain of inputs and outputs.

The key term is {\em independent}; we need mathematical machinery for dealing with it.
The Kolmogorov complexity $K(x)$ of a string $x$ is the length of the shortest program that produces $x$ on a universal Turing machine $U$:
\begin{equation} \label{def:kolmogorov}
K(x) = \min_p \{ |p| : U(p) = x \}.
\end{equation}
Roughly, $K(x)$ measures how nontrivial $x$ is. A simple pattern has low complexity, while a complex one has high complexity.
The conditional Kolmogorov complexity $K(x \mid y)$ is the length of the shortest program that produces $x$ given $y$ as auxiliary input:
\[ 
K(x \mid y) = \min_p \{ |p| : U(p, y) = x \}.
\]
Both quantities are defined up to an additive constant depending on the choice of $U$. Universality is not required for knowing. A universal Turing machine with no program has $N = 1$ and $\mathcal{I} = 0$; universality is a property of the interpreter, not of intelligence. The Kolmogorov complexity used in the independence condition is defined relative to a UTM, but this is merely a mathematical device to make $C$ well-defined.

$\mathcal{I}$ stands in dual relation to $K$: $K$ maps an output to its minimum program size, while $\mathcal{I}$ maps a program size to the maximum number of independent outputs. The $K$ direction has been developed for over half a century, including Rissanen's \citeyearpar{rissanen1978} Minimum Description Length; the complementary direction measured by $\mathcal{I}$ has not, to our knowledge, been articulated as a quantitative measure.

 The conditional $K(x \mid y)$ measures how much of $x$ is not already contained in $y$. When $K(x \mid y)$ is small, $y$ nearly determines $x$; when $K(x \mid y) \approx K(x)$, $y$ tells us nothing about $x$. This is the theoretically natural reference for independence.

\begin{definition}[Independent outputs] \label{def:independent}
Two outputs $o$ and $o'$ of a system $S$ are independent if neither can be predicted from the other by any description shorter than the output itself. Formally, 
\begin{equation}
 K(o \mid o') \approx K(o) \quad\text{and}\quad K(o' \mid o) \approx K(o').
\end{equation}
\end{definition}

The use of Kolmogorov complexity here has a natural operational interpretation. An interrogator who tries to predict $o$ from $o'$ is, in effect, searching for a short description of $o$ given $o'$. The most powerful conceivable interrogator, one with unlimited computational resources and optimal description, would find the shortest such description, achieving exactly $K(o \mid o')$. Kolmogorov complexity is therefore the {\em idealized interrogator}, the limit approached as the interrogator becomes more capable. A real interrogator is a computable approximation of this limit. The more intelligent the interrogator, the closer its judgment comes to $K$.

This reinterpretation has two consequences. First, it resolves the incomputability concern. $K$ is not a quantity to be computed in practice but a theoretical ideal that real interrogators approximate. The incomputability of $K$ is not a practical obstacle: we will shortly see (Section~\ref{sec:iteration}, Lemma~\ref{lem:iteration}) that iterative systems yield calculable order-level bounds on $K$. Second, it dissolves the worry about disagreement between interrogators. Two interrogators may differ, but they disagree only insofar as one is a better approximation of $K$ than the other. There is a fact of the matter, given by $K$, even if no finite interrogator reaches it.

For domains where such direct bounds are unavailable, alternative approximation formalisms exist. Section~\ref{sec:computability} discusses how the qualitative divergence criterion is robust to both kinds of approximation.

\begin{definition}[Knowing] \label{def:knowledge}
Let $n$ parameterize the size of domain $D$. A system $S$ {\em knows} $D$ if the number of independent outputs $N(S, n)$ grows without bound while $C(S)$ remains fixed, a growth rate achievable only by a system whose finite mechanism captures the generative structure of $D$. We write this condition as $\mathcal{I}(S, n) \to \infty$, since under these circumstances the intelligence density diverges as the domain grows.
\end{definition}

By domain $D$, we may take, e.g., the number of digits for arithmetic, the board size for chess, or the number of generated tokens for a language model (see Section~\ref{sec:fluency}). In each case $n$ indexes the number of times $F$ has been applied; for a system that iterates to produce output, the input-size parameter and the iteration-count parameter coincide.
Roughly speaking, $S$ {\em generalizes over} $D$ in this case, and we may use the two terms interchangeably throughout. The limit $n \to \infty$ is a convenient idealization; in practice, a sufficiently large $n$ suffices to distinguish knowing from memorization, as discussed in Section~\ref{sec:speed}.

\section{Classification of intelligence} \label{sec:classif}

With examples, we can categorize intelligence.\footnote{We use standard asymptotic notation throughout: $g(n) = O(f(n))$ means $g$ grows at most as fast as $f$ (upper bound); $g(n) = \Omega(f(n))$ means $g$ grows at least as fast as $f$ (lower bound); $g(n) = \Theta(f(n))$ means both hold simultaneously (tight bound). For example, $\log N = \Theta(n)$ for the multiplication algorithm means that $\log N$ grows exactly linearly with digit count.}

\subsection{No Computation, $\mathcal{I} \approx 0$}

A rock produces no distinguishable input-dependent outputs, so $N(S) \leq 1$, giving $\mathcal{I}(S) = 0$. More precisely, a rock's outputs (e.g., its temperature changes, vibration responses) are almost entirely predictable from one another, and the independence condition collapses the count to near zero. A river is filtered by the same condition. Its output (water flows downhill) is essentially one response to a vast range of inputs, since changing slope from $30^\circ$ to $31^\circ$ produces an output almost entirely predictable from the output at $30^\circ$. $N(S) \approx 1$, so $\mathcal{I} \approx 0$. That $\mathcal{I}$ assigns a near-zero value to rivers and rocks is not a weakness but a feature. Intelligence density, like temperature, is graded, and every physical system has some value of it, possibly very low.

\subsection{Memorization, $\mathcal{I} \to 0$} \label{sec:memo}

If a system stores every input-output pair explicitly, with $m$ entries of $b$ bits each, then $C = mb$, $N = m$, and $\mathcal{I} = \log_2 m / (mb)$, which is always less than $1$ and decreases as the table grows. A large lookup table has $\mathcal{I} \approx 0$. The system memorizes; it does not generalize.

A complementary challenge to the CRA is posed by Block \citeyearpar{block1981}, who imagined a ``Blockhead'': a giant lookup table storing correct responses for every possible conversation of bounded length. Blockhead passes the Turing test but clearly does not understand; this is the mirror image of Searle's argument. For unbounded domains, Blockhead is physically impossible (infinite storage required). For bounded domains, the challenge appears to survive. 

A Blockhead that stores one response for each possible 100-word conversation over a 10,000-word vocabulary has $N = 10^{400}$ entries. Each response is itself a 100-word string, roughly $1,300$ bits, so $C \approx 10^{403}$ bits. Therefore $\log_2 N \approx 1,300$ and $\mathcal{I} \approx 10^{-400}$, an intelligence density that is effectively zero. Longer conversations only worsen the ratio. The number of stored responses grows exponentially with length, while the information content of each response grows only linearly. Blockhead is a lookup table, and $\mathcal{I}$ diagnoses it as such regardless of how convincing its outputs appear.

The physical storage is prohibitive even for this modest case. The $\sim 10^{400}$ bits required exceed the number of atoms in the observable universe ($\sim 10^{80}$) by a factor of $10^{320}$. The word ``infinite'' is not required. A domain that is merely large enough to be linguistically interesting is already physically impossible for any lookup table.

\subsection{Computation without knowing, $\mathcal{I} = \Theta(1)$}

A single binary logic gate (XOR, AND, or OR) has two input bits, one output bit, and $C \approx 4$ bits of structure. It produces $N = 2$ independent outputs (0 and 1, neither predictable from the other). Therefore $\mathcal{I} = \log_2 2 / 4 = 0.25$. The gate computes, but its domain is fixed, with no $n$ to vary, no sense in which performance scales. $\mathcal{I}$ remains constant.

An $n$-bit adder circuit correctly adds any two $n$-bit numbers. One might expect this to count as knowing addition. Under our framework, it does not. The adder has $C = O(n)$: doubling the input width requires roughly twice as many gates. Since $N = 2^{2n}$ grows exponentially with $n$ but $\log N = 2n$ grows only linearly, $\mathcal{I} = \log N / C = \Theta(1)$. The intelligence density stays constant rather than diverging.

This is structurally analogous to the lookup table: in both cases, covering a larger domain requires more hardware. The adder is far more efficient ($C$ grows linearly with $n$ rather than exponentially) but neither achieves $\mathcal{I}(n) \to \infty$. The XOR gate and the $n$-bit adder both yield $\mathcal{I} = \Theta(1)$ by different routes (fixed domain versus growing domain with growing $C$) but the result is the same, namely computation without knowing.

This is a novel prediction of the framework. Two systems producing identical outputs over any finite domain are classified differently. The circuit computes addition correctly for its input width, but it does not {\em know} addition in the sense of Definition \ref{def:knowledge}, because knowing requires $C$ to stay fixed as $n$ grows. The algorithm knows because its finite description covers the infinite domain. This distinction corresponds precisely to the uniform/non-uniform divide in computational complexity theory \citep{arora2009}, a technical boundary long recognized but identified here with the clear distinction between knowing a domain and merely computing correct outputs on it.\footnote{In the advice string formulation, a non-uniform Turing machine receives an extra string $a_n$ for each input length $n$. The length of $a_n$ corresponds to the growth of $C$ beyond the fixed program, namely zero for uniform computation (knowing), $O(n)$ for linear-size circuits (category 3), and $O(2^n)$ for lookup tables (memorization). Shannon \citeyearpar{shannon1949} showed that almost all Boolean functions require circuits of size $\Theta(2^n/n)$, implying that most functions are not knowable. Knowing is the exception, not the rule.} Uniform computation uses a single Turing machine (fixed $C$) for all input sizes, while non-uniform computation uses a different circuit for each input size ($C$ grows with $n$). Non-uniform computation can even handle undecidable functions by hardwiring the answer for each $n$, the extreme case of memorization. Our four-way partition refines this classical dichotomy by further distinguishing the rate at which $C$ grows.

\subsection{Knowing, $\mathcal{I} \to \infty$}

Consider multiplication. A lookup table for single-digit multiplication stores $100$ entries of $\sim$7 bits each, giving $C \approx 700$ bits, $N = 100$, $\mathcal{I} = \log_2 100 / 700 \approx 0.01$. A 100$\times$100 lookup table stores $10,000$ entries of $\sim$7 bits each, giving $C \approx 70,000$ bits, $N = 10,000$, $\mathcal{I} \approx 0.00005$. 

Now consider the standard multiplication algorithm, which uses the 10$\times$10 lookup table plus $\sim$100 bits of carry and digit-decomposition rules, for a total of $C \approx 800$ bits. For multiplication of two $n$-digit numbers, there are $N \approx 10^{2n}$ distinct input pairs, so
\begin{equation}
 \log_2 N \approx 6.6n, \quad  \mathcal{I}(n) \approx \frac{6.6n}{800}.
\end{equation}
This produces all $10,000$ outputs of the 100$\times$100 table and infinitely more. The $9,900$ additional entries in the larger table are entirely redundant and derivable from the smaller mechanism. Storing them increases $C$ linearly with $N$ without increasing $\log N$, so $\mathcal{I}$ drops.

As the domain grows, $\mathcal{I}(n)$ diverges, growing without bound as $n \to \infty$, while the lookup table's $\mathcal{I}(n) \to 0$. The outputs are independent at the order level, by the argument given with Definition~\ref{def:independent}. This contrasts sharply with a pseudo-random number generator, where each output determines the next and $K(o' \mid o) \approx 0$. The precise value of $\mathcal{I}$ at any fixed $n$ depends on how $C$ is measured, but the divergence of $\mathcal{I}(n)$ is robust to any reasonable choice.

As a sanity check, an $n$-digit addition algorithm yields similar scaling: $\log_2 N \sim 6.6n$ with a smaller $C$ because no multiplication table is needed. The two algorithms have comparable $\mathcal{I}$, which is correct, since both generalize equally well across their respective domains. Multiplication is more {\em complex} (larger $C$), but not more {\em intelligent}. The metric correctly distinguishes complexity from intelligence.

\begin{table}[t]
\begin{center}
\begin{tabular}{lllll}
\hline \hline
System & $C(S)$& $\log_2 N$ & $\mathcal{I}$ scaling & Status \\ \hline
Rock & $> 0$ & $\approx 0$ & $\approx 0$ & No computation \\
River & $> 0$ & $\approx 0$ & $\approx 0$ & No independent outputs \\
Pseudo-random generator & $> 0$ & $\approx 0$ & $\approx 0$ & Outputs not independent \\
Lookup table ($n$ entries) & $\Theta(n)$ & $\Theta(n)$ & $\to 0$ & Memorization \\
$n$-bit adder circuit (family) & $O(n)$ & $\Theta(n)$ & $\Theta(1)$ & Computes, doesn't know \\
Binary logic gate & $O(1)$ & $\Theta(1)$ & $0.25$ & Fixed domain \\
Addition algorithm & $O(1)$ & $\Theta(n)$ & $\to \infty$ & Knows \\
Multiplication algorithm & $O(1)$ & $\Theta(n)$ & $\to \infty$ & Knows \\
Chess engine ($M$ squares) & $O(1)$ & $\Theta(M)$ & $\to \infty$ & Knows \\
LLM (multiple domains) & $O(1)$ & diverges & $\to \infty$ & Knows \\ [0pt]
Human brain & $O(1)$ & diverges & $\to \infty$ & Knows \\
\hline
\end{tabular}
\caption{$\mathcal{I}$ across the continuum of systems, showing $C(S)$, $\log_2 N$, and the resulting $\mathcal{I}$ scaling explicitly.
 A system {\em knows} its domain (Definition~\ref{def:knowledge}) if $\mathcal{I}(n) \to \infty$. $C = O(1)$ denotes a fixed deployed system. It can be large, but is independent of $n$ so that it does not grow with domain size.
 \label{t:summary}}
\end{center}
\end{table}

Note that the refutation of Blockhead, discussed in Section \ref{sec:memo}, does not require considering all $n$. A single sufficiently large $n$, one lying beyond Blockhead's storage range or at a domain intersection it does not cover, is enough to expose the failure. The interrogator need not probe the entire domain; one probe past the stored range suffices. This is the operational form of the practical test developed in Section~\ref{sec:speed}.

\subsection{An addition man}
\label{sec:adder}

To make the criterion concrete, we introduce a reference example used throughout the paper. Consider a man who can perform only one-digit addition with carry. He can perform no other operation directly, but he has a scratchpad.

Given $3 \times 4$, which he understands as addition of 3 four times, he proceeds by repeated addition. He writes a running sum and a counter on the scratchpad, adds $3$ to the running sum $0+3=3, 3+3=6,\dots$ and increments the counter at each step $1,2,\dots$, and halts when the counter reads $4$. The scratchpad then shows $12$. For $34 \times 27$, he runs a nested version of the same procedure with longer scratchpad usage, decomposing each two-digit number into digits, producing partial products by repeated addition, and accumulating them with positional shifts. The same rule handles every $n$; what grows is the number of applications and the scratchpad length.

A more familiar version of this man memorizes the multiplication table and the carry rules for multi-digit multiplication. This changes what is stored internally but not the anatomy of the computation: the multiplication table is a compact cache of results that single-digit addition would otherwise produce by repeated application, and carry handling is itself a small additional $F$ applied across digit positions. Whether the base table is memorized or reconstructed on demand, the fixed addition rule is applied iteratively to an external scratchpad. We return to this speed-versus-first-principles distinction in Section~\ref{sec:fluency}.

This example instantiates the anatomy to be developed next. The internal procedure $F$ is fixed with $C = O(1)$, the loop and the working memory are external (the pattern of scratchpad interaction, not a mechanism inside the man), and the two components $R$ (advance the running sum) and $H$ (counter reaches target) are both executed by $F$. The number of distinct products he can compute grows without bound with the digit count, so $\mathcal{I}(n) \to \infty$. The division of labor between internal $F$ and external scratchpad is the structure of every knowing system: transformers iterate fixed weights over an external context, brains apply fixed circuitry over external notations, the addition man applies fixed rules over paper. Subsequent references to ``the multiplication example'' refer to this setup.

\subsection{The meaning of intelligence density}
\label{sec:DD}

Table \ref{t:summary} summarizes the above and reveals a four-way partition. 
The intelligence criterion is $\mathcal{I}(n) \to \infty$, not the absolute value at any fixed $n$ (Definition~\ref{def:knowledge}). A bloated but correct implementation of an algorithm satisfies the criterion regardless of how low its density is at any single scale, because $C$ stays finite while $N$ grows. Conversely, a high-density system over a frozen domain (an XOR gate has $\mathcal{I} = 0.25$) does not know its domain in this sense, because extension is impossible.

Among systems that know, $\mathcal{I}$ measures how efficiently each uses its description length. Devoted machines have high density; general-purpose systems have low density because $C$ is large. The brain's $\sim$$10^{14}$ synaptic bits and an LLM's $\sim$$10^{12}$ parameters cover many domains with extensive redundancy. This low single-domain density is the cost of breadth, not a defect. Both kinds satisfy the divergence criterion; density distinguishes them at any fixed scale.

\section{Anatomy of a knowing system}

\subsection{The iteration principle}
\label{sec:iteration}

The difference between a lookup table and an algorithm reduces to {\em reuse}. A lookup table dedicates a unit of storage to each output and nothing is shared. An algorithm stores a rule $R$ once and applies it to every input. The carry rule for addition is stored once but used at every digit position, for every pair of numbers, at every scale. Reuse is a generalization, as discussed in Section \ref{sec:classif}. The mechanism that enables reuse is iteration.

A short iterative program, applied to inputs of growing size, forces output independence at the order level. The structural role of iteration can now be stated as follows. 
\begin{lemma}[Iteration]
\label{lem:iteration}
Let $S$ be an iterative system with transition function $F$ of description length $|F| = O(1)$, acting on inputs of size $n$. Let $o_i$ denote the output at step $i$, with $o_{i+1} = F(o_i)$ for $i = 0, 1, 2, \ldots$
\begin{enumerate}
\item \emph{(Induction.)} $K(o_{i+1} \mid o_i) \leq |F| + O(1)$. The next output is predictable from the previous one by one application of $F$.
\item \emph{(Independence.)} For typical pairs of outputs $o, o'$ on distinct inputs,
\begin{equation} \label{eq:iteration}
K(o \mid o') = \Omega(n).
\end{equation}
\item \emph{(Divergence.)} Consequently $\log N(S, n) = \Omega(n)$ and $\mathcal{I}(S, n) \to \infty$.
\end{enumerate}
\end{lemma}

\noindent\textit{Reasoning.} Part 1 follows from the definition of $F$: given $o_i$, applying $F$ once yields $o_{i+1}$. Describing this transition costs $|F|$ bits for $F$ itself plus a constant overhead to invoke it, so $K(o_{i+1} \mid o_i) \leq |F| + O(1)$.

For Parts 2 and 3, consider a typical\footnote{Compressible inputs can be exceptions, but they form a vanishing fraction, so when the full input domain of size $n$ is counted, the conclusion is unaffected \citep{livitanyi2019}.} input of size $n$. The output reached after successive applications of $F$ inherits the input's incompressibility: since $F$ is of fixed size, it cannot compress an input of complexity $\Omega(n)$ below $\Omega(n)$, and the resulting output $o$ has $K(o) = \Omega(n)$. For instance, if one factor in a multiplication is known, the missing information is proportional to the size of the remaining factor.

Two outputs $o$ and $o'$ produced on distinct typical inputs share only the transition function $F$; the data they carry come from different inputs and are therefore uncorrelated. Hence $K(o \mid o') \geq K(o) - |F| = \Omega(n) - O(1) = \Omega(n)$. Since the upper bound $K(o \mid o') = O(n)$ is established constructively (specify the input), independence holds at the order level, with $K(o \mid o') = \Theta(K(o))$. The typical inputs of size $n$ number $2^{\Omega(n)}$, so $\log N(n) = \Omega(n)$, and with $C = O(1)$, $\mathcal{I}(n) = \Omega(n) \to \infty$.

The induction and independence parts operate in opposite directions. Induction gives an \emph{upper} bound on the conditional complexity between adjacent outputs within a single successive chain: they are close, because one application of $F$ connects them. Independence gives a \emph{lower} bound on the conditional complexity between outputs belonging to different chains (produced on different inputs): they are far apart, because the input data differ. The same short $F$ produces both effects: it makes outputs predictable within a single computation, and it leaves outputs on different inputs uncorrelated across computations.

The converse of Lemma~\ref{lem:iteration} also holds. A system without iteration, that is a straight-line program of fixed description length, with no loop or recursion, can process inputs only up to size $O(C)$: each input read is a separate operation, and the number of operations is bounded by the program's source length. The number of distinct outputs is at most $2^{O(C)}$, finite, and $\mathcal{I}(S, n) = \Theta(1)$ rather than diverging. Iteration is therefore not merely sufficient but exactly the minimum requirement for knowing.

Lemma~\ref{lem:iteration} and its converse together give an equivalence:
\begin{equation} \label{eq:equivalence}
\mathcal{I}(S, n) \to \infty \iff S \text{ has a transition function } F \text{ with } |F| = O(1).
\end{equation}
The forward direction is Lemma~\ref{lem:iteration}; the reverse is the converse just stated, since iteration requires a transition function by Definition~\ref{def:transition}. The knowing criterion and the existence of a short $F$ are two expressions of the same condition.

\subsection{Externalization}

Working memory can be offloaded to external resources: paper for humans, tape for Turing machines, context windows for LLMs. In every case, the finite program that knows the domain is separated from the unbounded working space it consumes during execution. This externalization is compatible with the separation of machine description from tape in computability theory \citep{arora2009}, the extended mind thesis \citep{clark1998}, and the substrate independence of universal computation \citep{turing1950}. 

Humans illustrate this principle in its strongest form. Working memory in the brain is famously limited. Miller's \citeyearpar{miller1956} ``magical number seven'' bounds the items that can be held simultaneously, and the limit is even tighter for multi-step computation. The addition man of Section~\ref{sec:adder} is the canonical case: the rules he applies fit easily in memory, but the partial sums and counters he consumes during execution do not, so paper becomes necessary. The same person who multiplies reliably with a pencil fails without one, executing identical rules in both cases. Externalization is therefore not merely a convenience for humans but, for any computation of more than a few steps, a necessity, providing empirical confirmation that $C$ (the program) is properly separated from the working space.

$C(S)$ is the description length of the {\em entire system} that produces the outputs, including program, data, and whatever mechanism drives the repetition, and the system boundary is drawn around everything that contributes to the output.

\subsection{The transition function}

A general program has three structural parts: a working memory that holds intermediate state, a bounded loop that repeats a body of instructions, and a transition rule that computes the next state from the current one. For a knowing system, the first two are externalized. The working memory is the scratchpad, which lives outside the system's description as shown above. The loop is the pattern of repeated interaction between the system and the scratchpad, not a separate mechanism inside the system: the addition man of Section~\ref{sec:adder} does not hold a loop counter in his head, and an autoregressive language model does not contain one either; the inference engine feeds each output back as the next input. What remains inside the system is the transition rule alone.

This rule has two roles. It must advance the computation by one step, producing the next scratchpad state from the current one, and it must recognize when the goal has been reached, so that generation halts rather than continuing indefinitely. We therefore treat the two as components of a single function.

\begin{definition}[Transition function]
\label{def:transition}
The transition function $F$ of an iterative system is the finite rule that the system applies at each step. It contains two components: an advance component $R$ that computes the next scratchpad state from the current one, and a halt component $H$ that signals whether the goal has been reached. Both are internal to $F$; no separate loop mechanism is required.
\end{definition}

The components $R$ and $H$ are not architecturally distinct. In the addition man, $R$ is the rule that updates the running sum and $H$ is the check on the counter, executed as successive tests within the same single-digit-addition subroutine. In a transformer, a single forward pass computes both: the output distribution determines the next token ($R$) and may concentrate on the end-of-sequence token ($H$). The decomposition is functional, meaning that we can point to these two roles in the behavior of $F$, not that the system must realize them in separate modules.

By observing how the system produces the output $o_{n}$ from $o_{n-1}$, the interrogator can identify, directly or indirectly, the $F$ that the system applies. This identification bounds the domain-specific component of $C$. Once $F$ is exhibited, its size is a ceiling on how much of $C$ is devoted to this domain.

\subsection{Level of description} \label{sec:C}

The total description length $C$ depends on the chosen level of description. Even within a single digital system, measurement conventions can vary. A neural network's $C$ could be measured in 32 bits or 16 bits, yielding $\mathcal{I}$ values differing by a factor of two.

This ambiguity reflects scale-dependence, not conceptual weakness. The situation is analogous to effective theories in physics: at different energy scales, the relevant degrees of freedom change. Quantum chromodynamics describes quarks and gluons; nuclear physics describes protons and neutrons; chemistry describes atoms and bonds. Each level has its own variables, and quantities computed at one level need not equal those computed at another. No one considers this a deficiency of physics.

Similarly, $\mathcal{I}$ is computed within a chosen level of description, and different levels define \emph{different systems} with different degrees of freedom. The evaluation domain determines the appropriate level; once fixed, both $C$ and $N(S)$ are determined by the degrees of freedom at that level, and $\mathcal{I}$ is meaningful within that level. Constant-factor rescaling of $C$ affects snapshot values but not the divergence criterion, since it does not change asymptotic behavior. The core argument (Proposition \ref{prop:infinite}) requires only that $C$ is finite, which holds at any level for any physical system.

\section{Meaning as arrangement} \label{sec:arrangement}

Searle's Premise 3 \citep{searle1980},
\begin{equation} \label{Premise3}
 \text{Syntax is not sufficient for semantics,}
\end{equation}
assumes that meaning is something beyond the arrangement of symbols. The functionalist tradition in philosophy of mind \citep{putnam1967,fodor1975,dennett1987} established the converse for mental states: they are defined by their functional roles, not by their substrate. We extend this principle to meaning itself, offering a way to understand meaning over a domain $D$ as a selection and ordering of functions that produce correct outputs for all inputs from $D$. A complete account of meaning is neither possible nor intended here; we sketch only the core structure that bears on the framework of this paper, so we will be content with the following model that captures what matters here.

\subsection{Syntax is function composition}
\label{sec:syntax}

Syntax comprises exactly two elements: the {\em selection} of components (which to use) and the {\em ordering} of components (in what sequence to combine them). Consider examples:
\begin{quote}
The cat sat on the mat. \\
The cat sat on the coffee.\\
The mat sat on the cat.
\end{quote}
The first two sentences share the same arrangement but the second selects a word unfit for the slot: ``coffee'' does not satisfy the type ``sat on'' requires of its object. The failure is at the point of selection. In the first and the third, the same words are selected but the arrangement is altered: the third sentence places an inanimate object in the role of an animate sitter. The failure is at the point of ordering. Each failure is localized to a single decision in the composition; what determines meaning is selection and ordering together, not either alone.

Rules and operations are themselves compositions of lower-level operations: addition is a specific arrangement of XOR and AND gates, and even the choice of symbol set (binary, decimal, or Chinese numerals) is a selection of encoding functions. At the bottom, only basic gates and their arrangement remain. Syntax, fully analyzed, is function composition.

Modern language models realize this machinery empirically. The transformer mechanism described in Section~\ref{sec:composition} implements selection and ordering through a single next-word distribution. After ``The cat sat on the,'' the model assigns high probability to surfaces (``mat,'' ``chair,'' ``bench'') and negligible probability to ``coffee'' or ``idea.'' Selection is implemented as probability mass concentrated on type-appropriate words. Ordering is implemented in the same way. After ``The mat sat on the,'' the model assigns higher probability to surfaces and locations than to animate agents, because the preceding arrangement already fixed ``mat'' as subject. The same forward pass that produces fluent text on familiar inputs produces low probability on the violations above, without any module dedicated to type-checking or order-checking, since both fall out of the single learned distribution. The selection-and-ordering machinery posited above is what the model has learned to approximate, and the success of these approximations on held-out inputs is empirical evidence that the machinery is what natural language demands.

The identification of meaning with function composition has precedent across formal semantics \citep{montague1970,lambek1958}, the Curry-Howard correspondence \citep{curry1934,howard1980}, programming language theory \citep{scott1970,plotkin1981}, distributional linguistics \citep{firth1957,harris1954}, and connectionist and formal-systems traditions \citep{smolensky1988,hofstadter1979}.

\subsection{Tasks as function composition}
\label{sec:tasks}

Having identified syntax with function composition, we now show that correct function composition is what performing a task consists in. A task over a domain $D$ is a mapping from inputs to correct outputs. Any system that performs this task implements a composition of functions
\begin{equation}
 ( f_1, f_2, \dots, f_k )
\end{equation}
where the selection determines which functions are used and the ordering determines in what sequence they are applied. Meaning over $D$ is the correct selection and ordering, the arrangement that produces correct outputs for all inputs from $D$. We use ``function composition'' in the precise sense of mathematical composition over typed inputs and outputs; for symbolic and natural-language tasks the typing is implicit in selectional and structural constraints, as exhibited in Section~\ref{sec:syntax}.

The addition man of Section~\ref{sec:adder} illustrates this. His composition for multi-digit multiplication selects four functions: $f_1$ for digit decomposition, $f_2$ for base addition (the internal $F$), $f_3$ for carry propagation, and $f_4$ for accumulation. The composition $(f_1, f_2, f_3, f_4)$ is what we identify as knowing multiplication within this framework. Change the selection (replace $f_3$ with a different operation) and the outputs become wrong. Change the ordering (apply $f_4$ before $f_3$) and the outputs become wrong. The correct selection and ordering is the content of ``knowing multiplication'' as captured here.

The framework extends beyond arithmetic through a simple in-principle observation. A task whose correct outputs can be specified by a program is, by Church's thesis, a computable task; the program is its function composition written down. The class of such tasks is wide. A system that can read, write, and manipulate code therefore has, in principle, access to this entire class through code as the medium. Modern language models do operate over code at scale, which is one reason the framework's sketch is broadly applicable rather than confined to mathematical examples: the possibility of covering computable tasks via the code medium is open to any system that handles code, and the breadth of code-related capability in current language models is the empirical face of this possibility.

To see how this structure produces independence, return to the addition man's multiplication at large $n$. A single computation yields a sequence of intermediate outputs generated by iterating $F$: if $o_i = 756 \times 91$ is one output, then $o_{i+1} = F(o_i) = 756 \times 90$ is produced by the same $F$ applied to an adjacent input, and $K(o_{i+1}) \approx K(o_i)$, which is exactly what the interrogator expects, because the interrogator knows the algorithm. For any independently chosen $\hat{o}$, say $8238 \times 16385$ (a different input entirely), $K(o_i \mid \hat{o}) \approx K(o_i)$, meaning nothing about $\hat{o}$ predicts $o_i$. The iteration creates short-range dependence (successive outputs of $F$) while leaving cross-input independence intact. This is the structure that makes the system {\em know}. One $F$ covers many outputs, and distinct inputs remain distinct.

This is meaning produced by correct selection and ordering. Each output is exactly what the interrogator expects, since the correct composition was applied, yet it cannot be guessed from an unrelated output, so it is genuinely novel. Iteration delivers at each step a new meaningful output: meaning in the sense of Section~\ref{sec:syntax} and novelty in the sense of independence. The two together constitute knowing, and they are the two sides of Lemma~\ref{lem:iteration}: induction gives the short-range dependence that makes successive outputs correct under $F$; independence gives the distinctness of outputs on distinct inputs.

The same structure applies to any task that maps inputs to correct outputs. The multiplication example illustrates the general pattern: correct selection and ordering of functions is what performing the task consists in. PAC learning theory \citep{valiant1984,vapnik1998} formalizes one route to such systems: a hypothesis class of bounded VC dimension is learnable from finitely many samples, yielding a finite-$C$ system that generalizes across the distribution. PAC-learnability over an unbounded domain corresponds to the divergence $\mathcal{I}(n) \to \infty$, though PAC is a theory of learning (how $F$ is acquired from data, under iid assumptions) while $\mathcal{I}$ is a theory of evaluation (what a finite $F$ produces, without distributional assumptions).

We extend this line of reasoning beyond language, logic, and programs to {\em arbitrary tasks}. Any task that maps inputs to outputs is a function, and performing it correctly requires the right selection and ordering of sub-functions. Meaning over a domain is correct function composition over that domain.

Searle's Premise 3 in (\ref{Premise3}) is answered directly by the contextuality measure: correct function composition produces outputs with high $\mathcal{X}$ that are automatically independent of unrelated outputs (Corollary~\ref{cor:m-independent}). Where there is correct syntax in the sense developed here, that is the right selection and ordering of functions, the outputs that semantics is invoked to describe, correct and independent, are produced.

The identification is strongest for mathematical domains and must be transferred beyond them with care, as the following conditions make precise.

The extension remains valid wherever three conditions hold: (i) the domain can be specified well enough to separate correct from incorrect outputs, whether by domain structure if exact or by an evaluator if not; (ii) outputs admit a principled independence criterion, so that $N$ is not vacuously inflated by paraphrase; (iii) correctness does not require properties outside function composition, such as phenomenal experience. For natural-language tasks where an evaluator can adjudicate, such as question answering, translation, summarization, and code generation, these conditions obtain to the extent confirmed empirically by the success of language models.

The conditions can also fail in different ways. Domains whose correctness depends on a population of evaluators (aesthetic judgment, open-ended creative work, stylistic preference) admit evaluator-relative versions of the metric. When the evaluator distribution can itself be specified, for instance as preference data for trained reward models, these domains fall back within the framework, with the evaluator playing the role the domain structure plays for arithmetic; when no such specification is attempted, the metric becomes evaluator-relative along with the correctness it tracks. Domains where meaning is held to include phenomenal character lie outside our scope by construction (Section~\ref{sec:introduction}). The claim is therefore not that syntax captures every notion of meaning anyone has invoked, but that for any task-like domain where correctness is specifiable, whether directly or through an evaluator, meaning reduces to correct function composition.

\subsection{Composition paths, data, and independence}
\label{sec:composition}

The function composition view makes the source of output independence concrete. We illustrate it through transformer-based language models, which provide a clean realization of the structure.

\begin{definition}[Context]
\label{def:context}
The context at step $i$ of an iterative system $S$ is the sequence of all previously generated outputs:
\begin{equation} \label{context}
 x_i = (o_1, o_2, \ldots, o_{i-1}).
\end{equation}
The transition function $F$ acts on the full context: $o_i = F(x_i)$. The notation $o_{i+1} = F(o_i)$ used in Lemma~\ref{lem:iteration} is shorthand for this dependence on the entire prior sequence.
\end{definition}

A transformer processes a sequence of words\footnote{To be precise, tokens. A word can be further separated into sub-word tokens by the tokenizer; the framework applies at whichever granularity the system operates.} $x_i$ in (\ref{context}), where each $o_i$ is an output in the sense of Definition~\ref{def:context}. The autoregressive loop generates each successive word: given the current context $x_i$, a forward pass produces a probability distribution over the next word, from which $o_i$ is sampled, appended, and the process repeats. The same fixed weights apply at every iteration. This is the iteration mechanism of Section~\ref{sec:iteration}. The $F$ reapplied at each iteration is next-word prediction, that is, the forward pass itself, a fixed mapping from context to distribution over the next word.

The forward pass realizes function composition through the context. The effective function applied at position $i$ is
\begin{equation}
f_i = F(x_i),
\end{equation}
where $F$ is the same fixed forward pass at every position and $x_i$ is the context that varies. The fixed $f_j$ of Section~\ref{sec:tasks} becomes here a single trained forward pass $F$; what varies across iterations is not the function but the context that determines which routing within $F$ is effectively applied. The selection of Section~\ref{sec:syntax} is realized by the context (which determines what $F$ does), and the ordering is realized by the layered structure of $F$ together with the autoregressive loop. $F$ is reapplied until a termination condition is met, such as an end-of-sequence token or a length limit, and this repetition is itself a function of the context, so the entire procedure stays within function composition.

If working memory holds only a few items, how does $F$ process a context $x_i$ that may contain hundreds of prior outputs? $F$ does not read $x_i$ in its entirety. It selectively attends to the parts relevant to the current step and operates on that reduced representation. In a transformer, this selection is performed by the attention mechanism \citep{vaswani2017}, which computes a weighted sum over positions in $x_i$, where the weights are determined by the relevance of each position to the current step, and the result is a fixed-size vector that the feedforward layers then process. In humans, a similar role is played by chunking, whereby familiar patterns in the context are grouped into larger meaningful units \citep{chase1973}, so that $F$ operates on a manageable number of chunks rather than the raw sequence. The ability to form good chunks, that is to extract the relevant information from a long context, is itself part of what makes $F$ capable.

This realization exhibits a decomposition into {\em path} and {\em data}. The structural sequence of operations, the selection and ordering of functions that $F$ applies, is the path, encoded in the weights. The intermediate values flowing through the path are the data, carried by the residual stream. The path is shared across inputs within a domain (all multiplications follow the same steps), while the data varies per input (different digits produce different carries). Across domains, the paths themselves differ, producing independence. Quantitatively, the weights have fixed description length $C$ independent of problem size $n$, and the residual stream carries $O(n)$ bits of varying intermediate values. Since the shared part cannot reduce the order of the conditional, $K(o \mid o') = \Omega(n)$, recovering Lemma~\ref{lem:iteration}'s conclusion in the transformer-specific case.

\begin{theorem}[Context encodes $F$]
\label{thm:context}
Let $S$ be an iterative system with transition function $F$, and let $x_i $ be the context at step $i$ in (\ref{context}) with each $o_j = F(x_j)$. Then:
\begin{enumerate}
\item $K(o_i \mid x_i) \leq K(F \mid x_i) + O(\log n)$, and
\item $K(F \mid x_{i+1}) \leq K(F \mid x_i) + O(1)$.
\end{enumerate}
Consequently, $K(o_i \mid x_i)$ is monotonically non-increasing (up to the $O(\log n)$ application cost) as the context grows.
\end{theorem}

\noindent\textit{Reasoning.} Part 1 means that given $F$ and $x_i$, producing $o_i = F(x_i)$ requires only running $F$ on $x_i$ and specifying where in the output sequence we are, which costs $O(\log n)$ bits. Part 2 means that, since $x_{i+1} = x_i \cup \{o_i\} \supseteq x_i$, any program that computes $F$ from $x_i$ also computes $F$ from $x_{i+1}$ by simply discarding $o_i$ and running the same program, at a cost of $O(1)$ additional bits. So the sequence is non-increasing up to an additive $O(1)$ and is bounded above by $K(F \mid x_1) \leq |F|$. When $x_i$ is itself a trace of $F$, that is, when each $o_j$ in the context was produced by $F$ applied to $x_j$, the trace further constrains $F$: any candidate transition function $F'$ with $F'(x_j) \neq o_j$ for some $j \leq i - 1$ is inconsistent with the observed context and therefore excluded. For contexts long enough that this constraint substantially narrows the candidate set, $K(F \mid x_i) \ll |F|$, and consequently $K(o_i \mid x_i) \ll |F|$.

This bound is strictly stronger than the $K(o_{i+1} \mid o_i) \leq |F| + O(1)$ given by the Induction part of Lemma~\ref{lem:iteration}, which conditions on a single predecessor. The full context is what makes this possible: by reading it, the system recovers the structure of $F$ without a separate specification. This is also the information-theoretic content of externalized working memory (Section~\ref{sec:iteration}): the context serves as an external record from which the system reconstructs its own transition function at each step. The resulting measure, defined next, is graded by construction.

\begin{definition}[Contextuality]
\label{def:contextuality}
The contextuality of output $o_i$ given context $x_i$ is
\begin{equation} \label{contextuality_eq}
\mathcal{X}(o_i \mid x_i) = \frac{1}{K(o_i \mid x_i)}.
\end{equation}
\end{definition}

An output has higher contextuality when it is more predictable from its context, that is when the context already contains the structure needed to produce it. Intuitively, we can guess the next term of a series more easily if more initial terms are given; we can guess the next word of a sentence more easily if more context is given. Each prior output constrains what follows. By Theorem~\ref{thm:context}, outputs of an iterative system with a short $F$ have high contextuality: $\mathcal{X}(o_i \mid x_i) \gg 1/|F|$. Conversely, an unrelated output $\hat{o}$ carries no information about $o_i$ and leaves $K(o_i \mid \hat{o})$ unchanged.\footnote{Note that $K(o_i \mid x_i)$ measures information content, not access cost: $x_i$ is available to the program, but nothing requires the program to read all of $x_i$. The selective access discussed in Section~\ref{sec:iteration}, attention in transformers and chunking \citep{miller1956,chase1973} in humans, is itself part of the shortest program that produces $o_i$ from $x_i$.} The following corollary makes this precise.

\begin{corollary}[Contextuality implies independence]
\label{cor:m-independent}
Let $o_i$ be an output with $K(o_i \mid x_i) = \epsilon$ and $K(o_i) = \Omega(n)$. Let $\hat{o}$ be an output produced on an input unrelated to $x_i$, so that $K(x_i \mid \hat{o}) \approx K(x_i)$. Then
\[
K(o_i \mid \hat{o}) \geq K(o_i) - O(\epsilon).
\]
\end{corollary}

\noindent\textit{Reasoning.} Since $o_i$ is determined to within $\epsilon$ bits by $x_i$, any description of $o_i$ from $\hat{o}$ must first recover $x_i$ to within $\epsilon$ bits. But $\hat{o}$ is unrelated to $x_i$, so $K(x_i \mid \hat{o}) \approx K(x_i)$: knowing $\hat{o}$ does not help reconstruct $x_i$. Therefore $K(o_i \mid \hat{o}) \geq K(o_i) - O(\epsilon)$, and since $\epsilon \ll K(o_i)$, the output $o_i$ is independent of $\hat{o}$.

High contextuality therefore subsumes independence: an output that is nearly determined by its context cannot be predicted from an unrelated output. The two conditions of Lemma~\ref{lem:iteration}, induction (predictability from the immediate predecessor) and independence (unpredictability from unrelated outputs), are unified by $\mathcal{X}$. High $\mathcal{X}$ is the single condition from which both follow.

The unification has a concrete mechanistic basis. As noted in Section~\ref{sec:iteration}, the transition function $F$ decomposes into ``compare the scratchpad state with the goal'' $H$ and the iterative relation $R$. When $H$ binds to {\em this} context $x_i$, it simultaneously (i) determines the correct output for this context with high $\mathcal{X}$ and (ii) fails to bind to any unrelated context (independence).\footnote{In a transformer, this binding is realized by the Q$\cdot$K attention mechanism: the query encodes ``does this context match the goal?'' and the keys encode the scratchpad state.} The same comparison operation that produces correctness produces independence. Binding to one scratchpad is unbinding from all others. This is why $\mathcal{X}$ subsumes both. The comparison that makes the output predictable from its own context is exactly what makes it unpredictable from an unrelated one.

Theorem~\ref{thm:context} and Corollary~\ref{cor:m-independent} together predict a precise two-case structure. Theorem~\ref{thm:context} assumes the context $x_i$ is a trace of the {\em same} transition function $F$. When this holds, that is when the context is about the same topic, task, or domain, longer context encodes $F$ more completely, $K(o_i \mid x_i)$ decreases, and outputs become more predictable, with higher contextuality. When the topic changes, a new transition function $F'$ begins. The prior context, a trace of $F$, does not encode $F'$, so the new output has low contextuality {\em relative to the old context}, and by Corollary~\ref{cor:m-independent} is independent of the outputs under $F$. An LLM given a sequence of multiplications followed by ``What is the capital of France?'' instantiates exactly this structure: within the arithmetic subsequence, successive outputs encode the addition man's composition (Section~\ref{sec:adder}) in the transformer's learned weights, and $\mathcal{X}$ rises; when the domain shifts to geography, the prior context does not encode the geography $F'$, and the new output is independent of the arithmetic context. The framework thus distinguishes two sources of apparent unpredictability: insufficient context within a single domain, resolved by extending the context under the same $F$, and a genuine change of domain, which introduces a new $F'$ and correctly yields independence. Conflating the two, that is treating all long-context degradation as an engineering failure, misdiagnoses the situation.

Confirming intelligence in practice does not require observing $\mathcal{I}$ itself diverge. It suffices to observe outputs with high contextuality $\mathcal{X}$ in (\ref{contextuality_eq}), independence from unrelated inputs (Definition~\ref{def:independent}), and appropriate termination: these witness the transition function $F$ operating over a scratchpad (Section~\ref{sec:iteration}), and by the equivalence (\ref{eq:equivalence}), the existence of $F$ over a growing domain entails $\mathcal{I}(n) \to \infty$. For an LLM, $n$ counts generated tokens (see Section~\ref{sec:fluency}), and the test is: as $n$ grows, observe whether successive tokens have high $\mathcal{X}$ (correct), whether outputs on unrelated inputs remain unpredictable from one another (independent), and whether generation terminates appropriately. If all three hold as $n$ increases, the model knows its domain.

The mechanism that separates meaning (path) from independence (data) is the learned weights themselves. Identical underlying composition paths produce identical internal activations across forward passes; differing data produce different residual values. Training shapes the weights so that this factoring happens automatically: shared paths get encoded in weights during gradient descent, while data variations are routed through the residual stream during inference.

When the system's domain is not known in advance, the interrogator discovers it by probing different domains and estimating the independent output count $N_i$ in each. Since outputs from genuinely different domains are independent of one another (knowing $123 \times 456$ does not predict the best chess move), the total count decomposes additively: $\log N_{\text{total}} = \sum_i \log N_i$. This is the formal content of what a Turing test does.

\section{Operationalizing the criterion}
\label{sec:operationalizing}

\subsection{Correctness is relative}

The definition of $\cal I$ makes no reference to correctness.\footnote{Throughout this paper ``correct'' is shorthand for ``correct relative to the specified domain.''} $\mathcal{I}$ measures generative capacity; correctness is a separate question determined by the domain's structure. A system with $\mathcal{I}(n) \to \infty$ that produces wrong outputs is a powerful but poorly calibrated generalizer, not an unintelligent one. This separation is what makes $\mathcal{I}$ observer-independent. Correctness is {\em always} relative to a domain. $1+1=0$ is correct in a field with characteristic 2; a medical diagnosis is correct relative to the patient's condition; a conversational response is correct relative to an evaluator's judgment. The same output might be called creative in one context and erroneous in another; creativity and hallucination are the same generative capacity, judged by different evaluators.

This domain-relativity also answers classical objections to identifying syntax with semantics \citep{kripke1982,putnam1975,fodor1975}. Each assumes that meaning should be specifiable from the system alone. Our framework rejects this. Meaning is correct function composition over a {\em specified} domain; once the domain is fixed, the ambiguities these objections raise dissolve. The arrangement plus the domain constitutes meaning; neither alone does.


Many nontrivial properties follow from knowing and iteration.

\subsection{Interrogator}

Independence, as defined in Definition~\ref{def:independent}, is binary, but in practice it is graded: ``A dog barks'' and ``A dog runs'' share the subject ``dog'' and are neither fully independent nor fully dependent. A continuous measure based on the normalized conditional Kolmogorov complexity $K(o_1 \mid o_2)/K(o_1)$, equivalent to the normalized information distance of \citet{li2004} applied to outputs, captures this gradation and yields a continuous effective output count.

This gradedness is interrogator-dependent. Consider ``a lovely dog'' and ``an adorable dog.'' A generic compressor finds these outputs nearly identical (shared frame, near-synonymous adjectives) so $K(o \mid o')$ appears small and the outputs appear dependent. A poet with a high-resolution vocabulary for aesthetic modifiers registers ``lovely'' and ``adorable'' as carrying distinct, non-redundant information, perhaps as distinct as a geographer finds ``a river'' and ``a mountain,'' while the same poet may lack the vocabulary to separate river and mountain as reliably. Each observer approximates $K$ with finer resolution in its own domain of expertise: expertise is the capacity to detect distinctions that others miss.

A more refined scale does not remove this variability; it only relocates it. The idealized interrogator of Definition~\ref{def:independent}, the limit of approximating $K$ exactly, detects the distinction whenever one exists in principle, returning the judgment to binary. Graded dependence is therefore an artifact of real interrogators' finite resolution rather than a property of the outputs themselves. The genuine observer-independent claim is qualitative, whether the system exhibits a transition function $F$ that operates for sufficiently large $n$. This verification takes different forms in different domains: direct physical readout at small $n$ in sensor-anchored domains (robot collisions, temperature readings), and learned $H$ patterns at higher $n$ in abstract domains (a mathematician recognizing a valid inference, a native speaker detecting ungrammaticality). Order of growth is robust to the evaluator's resolution (constant-factor differences in detection do not change divergence behavior) and to the choice of binary or graded independence. We therefore do not develop a formal graded calculus in this paper: the qualitative criterion carries the theoretical weight.

\subsection{Computability}
\label{sec:computability}

$\mathcal{I}$ depends on Kolmogorov complexity $K$, which is provably incomputable \citep{kolmogorov1965}, and lower bounds on $K$ are in general unattainable: Chaitin's incompleteness theorem \citep{chaitin1974} shows that no formal system can prove $K(x) \geq c$ for sufficiently large $c$. Lemma~\ref{lem:iteration} provides a structural bypass. Because the shared structure (the transition function $F$) is known to be of fixed size $O(1)$, for typical inputs the residual (the data) must be $\Omega(n)$. The upper bound on shared structure implies a lower bound on independence in the typical case, which is all that the divergence of $\mathcal{I}(n)$ requires. The divergence for iterative systems is therefore not merely conjectured but provable at the order level, a rare situation in which the incomputability of $K$ does not obstruct the conclusion. In practice, the determination reduces to checking the existence of the transition function $F$ for sufficiently large $n$, at the resolution available to the interrogator.

\section{Objections and tests}

\subsection{Why fluency appears intelligent}
\label{sec:fluency}
\label{sec:speed}

The iteration principle requires that a short program $F$ be applied repeatedly across a growing domain. Yet most intelligent systems observed in practice, including LLMs and human brains, answer typical questions almost instantly, without running long loops. A system that took minutes to multiply $3 \times 4$ would seem less intelligent than one that answers in a second, even though the first runs the full iteration from base principles while the second does not. The appearance of intelligence tracks response speed, and response speed is achieved by {\em memoization}, that is, storing intermediate results and starting iteration from a cached state rather than from scratch.

A child learning single-digit multiplication does not, in principle, need to loop. The product $3 \times 4$ can be computed by adding $3$ four times. However, we memorize the multiplication table, so the same answer comes out in one lookup step instead of four iteration steps. This caching is not merely convenient but necessary: the working-memory limit of Section~\ref{sec:iteration} rules out performing even moderately long iterations from scratch. Two-digit multiplication then runs a few iteration steps over these memorized base facts, rather than many additions from scratch. The iteration mechanism is still there (each word still requires a forward pass) but the loop that traverses stored structure is short.

Answering a natural-language question works the same way. Intermediate results and frequent patterns are stored; the system starts from a cached state and runs a few steps to produce an answer. This is why explanation does not recede into infinite regress: it bottoms out in shared background knowledge from which short chains of inference reach the surface.

Within the framework, memoization trades density for speed \citep{dennett1987}. Storing cached results increases $C$ without proportionally increasing $N$, so $\mathcal{I}$ decreases, but response time decreases much more. This is the mechanism behind the density--speed tradeoff noted in Section \ref{sec:DD}. A bare iterative program has the highest $\mathcal{I}$ per bit but may be slow; a system with extensive memoization has lower $\mathcal{I}$ due to the larger $C$ but answers instantly. Both satisfy the knowing criterion. What the iteration principle requires is that a short $R$ and a termination condition $H$ remain in the system; how much iteration is short-circuited by stored intermediate results is a separate choice, governed by the speed demands of the domain.

This reconciles an apparent tension. Humans typically answer in two or three steps, not by running the full iteration from first principles, yet the knowing criterion requires that $F$ operate across the domain as it scales. The criterion refers to the system's capacity across the domain, not to the computational depth of any single query. What matters is that the system produces $N(n)$ independent outputs as $n$ grows, using a fixed program $C$; how much of each output is recovered from cached intermediate results versus recomputed from the base iteration is a separate question about speed. For unfamiliar inputs, the bare iteration is recoverable. A person who has memorized the multiplication table can still work out $743 \times 862$ with pencil and paper, running essentially the algorithm of Section~\ref{sec:iteration}.

This also suggests a practical test for knowing. By the equivalence (\ref{eq:equivalence}), checking $\mathcal{I}(n) \to \infty$ reduces to exhibiting $F$. An interrogator selects a sufficiently large $n$, beyond the range where memoization is plausible, and examines whether the system produces $o_n$ from $o_{n-1}$ by a short, regular transition. If it does, $F$ is present; if the output bears no systematic relation to its predecessor, the system was exploiting stored entries that ran out. A more capable interrogator recognizes $R$ across a wider range of domains and closes off more memoization strategies, paralleling the role of the interrogator in the independence condition.

For multiplication, $n$ is the digit count and is read directly from the input. For a language model, $n$ is the number of generated tokens, that is the number of times $F$ has been applied to the scratchpad. Unlike multiplication, the minimal $n$ required to answer a given question cannot be determined in advance. The interrogator cannot specify beforehand how many tokens ``Explain quantum entanglement'' requires. However, verification remains possible through the two components of $F$. The interrogator checks $R$ by examining whether each successive token follows from the context by a legitimate transition, and checks $H$ by examining whether the system terminates at an appropriate point. What matters is not knowing $n$ in advance but confirming, after the fact, that $R$ connected $o_{i-1}$ to $o_i$ at each step and that $H$ fired at the right moment.

A reason for a large $C$ is multi-domain coverage: a system that handles many domains simultaneously requires a larger mechanism than one specialized to a single domain, even if each domain's function composition is individually compact. Biological brains and large language models both exhibit large $C$ for both reasons, speed optimization and multi-domain coverage, which is why their $\mathcal{I}$ in any single fixed domain is modest. The qualitative determination that $\mathcal{I}$ does not vanish as the domain scales is unaffected by either tradeoff.

\subsection{Thinking and explaining}
\label{sec:thinking}

An LLM generates text by predicting the next token from a deterministic $F$. When asked why it gave an answer $o$, it does not produce a trace of its internal activations but a content-level justification. The criticism that LLMs ``cannot explain themselves'' rests on identifying explanation with mechanism-tracing, that is, reporting the activation chain that produced the output.

But this is not what ``why did you say that?'' usually asks. What we want is a causal or logical justification at the level of content, that is, reasons that, if true, make the answer make sense. An LLM can produce such justifications, since the explanation is itself an output $F(x_i)$ conditioned on the context $x_i$ that includes the original answer, and when its training has been consistent with the domain, the justifications are correct in the same sense any causal explanation is correct, namely consistent with the structure of the domain. This second kind of explanation is itself produced by the same forward pass, through another round of next-token prediction conditioned on ``why?''; its content is a domain-level story, not a mechanism-level trace.

The structure seems the same in human cognition. When a person explains why they said something, they do not report neural activations; they produce a content-level account. The brain's mechanism is as inaccessible to the explanation-generating process as the LLM's forward pass is to its justification-generating forward pass. Both systems explain at the content level; neither has transparent access to its own mechanism.

The same observation bears on the symbol grounding problem \citep{harnad1990}, which holds that LLMs manipulate symbols without ever connecting them to the world and therefore lack meaning. Whatever grounding a speaker may possess, a text produced by a grounded speaker and a text produced by an ungrounded one can be identical as strings. Grounding is a property of the producer, not of the string; it does not survive into the written form. Communication proceeds through strings, and so whatever the framework calls ``meaning'' must be readable from the string itself, as the correctness of function composition. That a speaker saw a dog before writing ``the dog barked'' leaves no residue in the text that another, text-trained, speaker could fail to leave.

Within the framework, the capacity to explain is not a separate faculty but an extension of the domain: the system produces not only answers but also justifications, and these outputs must satisfy the same criteria of independence and correctness. When they do, $F$ covers the joint domain of answers and justifications. Explainability is the knowing criterion applied to the domain of self-justification.

Recent reasoning models, trained to generate extended chains of thought \citep{wei2022} before final answers, illustrate this framing concretely. Their architecture is unchanged from standard language models, namely the same autoregressive loop, with next-token prediction as $F$. What differs is training. When the content requires causal or logical connection between sentences, the model has been trained to produce an explicit chain of thought rather than jump to the conclusion. The working-out appears in the token stream because training rewards its presence, not because any new mechanism has been added. In framework terms, reasoning models are the same knowing systems as standard language models with the justification domain made explicit; $\mathcal{I}$ over the joint domain of answer and chain of thought captures the denser self-explanatory content they produce.\footnote{The content-level versus mechanism-level distinction appears to align with a familiar distinction in philosophy of mind: consciousness seems to operate at the content level, unconscious processes at the mechanism level. Whether this structural parallel reflects a deeper similarity is beyond the scope of this paper.}

\subsection{Falsifiability}

{\em False positive:} a system with $\mathcal{I}(n) \to \infty$ that no one would call intelligent. A pseudo-random number generator has deterministic structure: $K(o' \mid o) \ll K(o')$, and the independence condition filters it out. A true random source satisfies the criterion technically, a boundary set by Shannon's source coding theorem \citep{shannon1948} that applies to any observation-based measure; correctness (Section~\ref{sec:arrangement}) is the operative criterion for whether generative capacity is directed at a domain.

{\em False negative:} a system with $\mathcal{I}(n) \to 0$ that everyone would call intelligent. This would require a system that clearly knows a domain but achieves its performance through pure memorization, a lookup table that knows. Block's Blockhead is the canonical thought experiment here. For unbounded domains it is physically unrealizable (Proposition \ref{prop:infinite}). For bounded but linguistically interesting domains, Section~\ref{sec:classif} establishes two further reasons Blockhead fails: (i) it cannot answer questions at the intersection of domains it has not explicitly stored, and (ii) the storage required for realistic conversational coverage exceeds the physical capacity of the universe. In all these cases, $\mathcal{I}(n) \to 0$ correctly diagnoses memorization.

{\em Framework commitment on externalization.} The externalization principle of Section~\ref{sec:iteration} carries its own refutation condition. A demonstration that some form of intelligence requires working memory that cannot in principle be externalized would refute the principle and require revising the separation between $C$ and $W(n)$ assumed throughout.

An empirically accessible test is to measure $\mathcal{I}$ for a range of systems and compare with human judgments of intelligence. We consider this an important direction for future experimental work.

\subsection{On Putnam's triviality argument}

Putnam \citeyearpar{putnam1988} argued that every ordinary open physical system implements every abstract finite state automaton, making computational descriptions trivial. If this applied to our framework, every system would have $\mathcal{I} = \infty$.

The independence condition blocks this. Putnam's construction works by {\em relabeling} physical states post hoc. But relabeling does not create genuinely independent outputs. If dynamics at $t_1$ determine dynamics at $t_2$, then $K(o_{t_2} \mid o_{t_1}) \ll K(o_{t_2})$, and the outputs are not independent. A wall's molecular motion produces no genuinely independent outputs in response to distinct inputs, so $N(S) \approx 0$ and $\mathcal{I} \approx 0$.

\section{Scope and philosophical position}

\subsection{The Redefinition Objection}

Perhaps the most fundamental objection is that we have redefined intelligence, not shown that machines have it. This presupposes a ``real'' intelligence that our definition fails to capture. No consensus definition exists \citep{legg2007}. Our $\mathcal{I}$ completes Turing's move: if intelligence exists as a natural kind, $\mathcal{I}$ measures it; if it does not, $\mathcal{I}$ still measures a real physical quantity that does explanatory work. The use of Kolmogorov complexity is standard in the algorithmic information theory tradition \citep{solomonoff1964,hutter2005,schmidhuber2003}. As for the objection that intelligence requires goals: $\mathcal{I}$ measures the computational capacity that {\em enables} goal-achievement, without requiring goals to be specified.

The above model captures what people already do. When judging intelligence, humans implicitly estimate intelligence density in three steps: (1) a {\em capacity filter}, dismissing systems with obviously small $C$ (rivers, rocks) as candidates; (2) {\em output diversity}, that is how many different, correct responses across different situations; and (3) {\em independence detection}, recognizing when an output is merely a rephrasing versus genuinely new information, an informal approximation of $K(o' \mid o) \approx K(o')$. The summarizing Table~\ref{t:summary} confirms that $\mathcal{I}$ reproduces these intuitive judgments across the full continuum from rocks to brains.

Three subsidiary objections can be addressed briefly. The observer-dependence objection \citep{searle1992} is answered by the independence condition, which is a fact about $K$, not about observers, and by $C$ in the denominator, which distinguishes a lookup table from an algorithm even when their outputs match. The qualia objection \citep{nagel1974,chalmers1995} conflates intelligence with consciousness; this paper defines only the former, and Integrated Information Theory \citep{tononi2004} targets the latter through a structurally similar information-theoretic measure. The stochastic parrot objection \citep{bender2020} is answered by the contextuality measure: outputs with high $\mathcal{X}$ are correct relative to their context and independent of unrelated outputs (Corollary~\ref{cor:m-independent}), the opposite of recombination.

\subsection{Intentionality, originality and scope}
\label{sec:intentionality}

We close by gathering the framework's position on intentionality and originality, and stating what the framework supports.

On the present account, intentionality is the existence of $F$, the finite generative structure that produces correct and independent outputs over a domain. It is not an additional property over and above $F$, nor is it constituted by the interrogator's judgment; rather, the interrogator detects it. This places the account within the deflationary functionalist tradition \citep{dennett1987}: the operational condition is that a sufficiently capable interrogator identify $F$ from the system's outputs, and the idealization of this interrogator (Definition~\ref{def:independent}) recovers $F$'s existence as the observer-independent fact being approximated. Current trained language models satisfy this detection condition in the sense that human interrogators do identify their transition functions from outputs.

The identification is not stipulative but a rereading of the equivalence (\ref{eq:equivalence}): knowing and $F$'s existence are the same condition, so attaching intentionality to one is attaching it to the other.

This is a competing frame rather than a refutation of Searle's position on its own terms. The distinction between original and derived intentionality is rejected symmetrically: all intentionality is the existence of $F$, whether realized in silicon or in biological tissue. The substrate independence established earlier applies to intentionality as much as to intelligence, and the asymmetry Searle draws between machine and brain is not available within this framework.

This paper addresses the {\em state of knowing} for a given $F$: whether a fixed finite arrangement generalizes over a domain, and whether its outputs are correct and independent. The paper does {\em not} address how $F$ came to be (whether by a programmer, by gradient descent, or by evolution), how a system acquires or modifies $F$ for new domains, or what happens at the mechanism level beneath $F$'s observable behavior.

The addition man of Section~\ref{sec:adder} illustrates the scope precisely: his $F$ produces correct outputs, and the framework tells us he knows multiplication; it does not tell us how he came to possess that $F$ (whether memorized, taught, or reinvented), nor whether acquiring it was itself an act of understanding. Current trained language models occupy the same structural position: $F$ is identifiable from their outputs and the outputs are judged correct by competent interrogators, while the training trajectory that arrived at $F$ is a separate matter.

This mapping aligns with Searle's own premise structure. Premise 3, that syntax is insufficient for semantics, is refuted within the scope defined here: where correctness is specifiable, correct function composition produces outputs with high $\mathcal{X}$ that are automatically independent of unrelated outputs (Section~\ref{sec:arrangement}, Corollary~\ref{cor:m-independent}). Premise 2, that meaning requires an intentionality not exhausted by arrangement, is not refuted; it is reframed, with intentionality identified as $F$'s existence, and its residual force relocated to the origin of $F$, which belongs to understanding rather than knowing. The engagement with the Chinese Room is therefore scoped by design: Premise 3 fails within knowing, while whatever Premise 2 retains beyond this reframing awaits the treatment of understanding.

The distinction between content level and mechanism level developed in Section~\ref{sec:thinking} for explanation applies here equally: content-level aboutness of an output is itself an output of $F$ on context extended to include such a question, and is verifiable by the same independence and correctness criteria that govern any $F$ output, and is therefore inside the scope; mechanism-level aboutness is outside, in the same sense and for the same reason that mechanism-level explanation is outside.

What remains genuinely outside, not because mechanism is demanded but because the question is different, is the construction and modification of $F$ itself. We identify this with \emph{understanding} rather than knowing, and a companion framework is left to future work. The framework shows that {\em knowing can be isolated from understanding.} Legg and Hutter's universal intelligence measure \citep{legg2007}, which averages an agent's performance over environments weighted by their Kolmogorov complexity, fits naturally into this division as a measure of understanding rather than a competitor: it quantifies the capacity to perform across environments, which in our terms is the capacity to supply or adapt an $F$ for each, rather than the capacity of a fixed $F$ within one domain. On this reading $\mathcal{I}$ and the Legg--Hutter measure are complementary.

\section{Conclusion}

Prior debates about machine intelligence have been qualitative, asking whether the system thinks or not. This paper replaces the binary question with a quantitative measure, $\mathcal{I}(S) = \log N(S) / C$, and shows that the qualitative distinction between knowing and not knowing emerges from its asymptotic behavior rather than from an imposed threshold. The measure places intelligence on a substrate-independent continuum, requires no observer, no externally defined reward, and no environment specification, and distinguishes generalization from memorization via the independence condition.

The contextuality measure $\mathcal{X}$ complements $\mathcal{I}$ by evaluating what the system produces. It unifies correctness and independence into a single condition and thereby offers a way to understand how correct function composition produces what semantics is invoked to explain. Longer context within the same domain increases $\mathcal{X}$, while a change of domain correctly yields independence, distinguishing engineering limitations from the natural structure of multi-domain conversation.

The entire framework reduces to a single element: the transition function. Everything else follows. Its iteration means intelligence; its trace in the context means contextuality; its binding to the context gives independence; its comparison of state to goal gives termination and correctness. Intelligence, meaning, independence, and correctness are not four separate requirements but four aspects of a single structure: a finite transition function applied to a scratchpad.

This paper's claims are restricted to the state of knowing. The origin of the transition function, the construction and modification of arrangements, and whatever remains of intentionality beyond its detectability, are the province of \emph{understanding}, and are left to future work.

\section*{Acknowledgments}
The author thanks the students of Science and Civilization (since 2012) for helpful discussions. He used Claude (Anthropic) for editing and discussion during the manuscript preparation. The final content was reviewed and approved by the author, who takes full responsibility for the work.

\bibliographystyle{plainnat}

\end{document}